\def\year{2021}\relax
\documentclass[letterpaper]{article} % DO NOT CHANGE THIS
\usepackage{aaai21}  % DO NOT CHANGE THIS
\usepackage{times}  % DO NOT CHANGE THIS
\usepackage{helvet} % DO NOT CHANGE THIS
\usepackage{courier}  % DO NOT CHANGE THIS
\usepackage[hyphens]{url}  % DO NOT CHANGE THIS
\usepackage{graphicx} % DO NOT CHANGE THIS
\usepackage{natbib}  % DO NOT CHANGE THIS AND DO NOT ADD ANY OPTIONS TO IT
\usepackage{caption} % DO NOT CHANGE THIS AND DO NOT ADD ANY OPTIONS TO IT
\usepackage{graphicx}
\usepackage{algorithm}  
\usepackage{algpseudocode}  
\usepackage{amsmath}
\usepackage{booktabs}
\usepackage{amsfonts}
\usepackage{subfigure}
\usepackage{makecell}

\title{A Traffic Light Dynamic Control Algorithm with Deep Reinforcement Learning Based on GNN Prediction}

\author {
		Xiaorong Hu,\textsuperscript{\rm 1}
      Chenguang Zhao,\textsuperscript{\rm 1}
	Gang Wang \textsuperscript{\rm 1} \\
}
\affiliations {
	\textsuperscript{\rm 1} Department of Electronic and Information Engineering, Beihang University \\
	\{hxiaorong,zchenguang,gwang\}@buaa.edu.cn
}

\begin{document}

\maketitle

\begin{abstract}
Today's intelligent traffic light control system is based on the current road traffic conditions for traffic regulation. However, these approaches cannot exploit the future traffic information in advance. In this paper, we propose GPlight, a deep reinforcement learning (DRL) algorithm integrated with graph neural network (GNN) , to relieve the traffic congestion for multi-intersection intelligent traffic control system. In GPlight, the graph neural network (GNN) is first used to predict the future short-term traffic flow at the intersections. Then, the results of traffic flow prediction are used in traffic light control, and the agent combines the predicted results with the observed current traffic conditions to dynamically control the phase and duration of the traffic lights at the intersection. Experiments on both synthetic and two real-world data-sets of Hangzhou and New-York verify the effectiveness and rationality of the GPlight algorithm.
\end{abstract}
\section{Introduction}

\noindent With the rapid increase of vehicle quantity, traffic congestion has become an urgent problem to be solved in many places around the world, especially in big cities. In order to solve this problem, reduce the waiting time of vehicles on the road and increase the carrying capacity of urban road network, Intelligent Transportation System (ITS) has become one of the hot research issues in recent years, which aims to optimize the coordination and control of traffic flow.

Traffic light control is an important part of ITS. In recent years, reinforcement learning (RL) technology has become one of the most widely used approaches and has been extensively used for traffic light control problem\cite{1wei2019survey}. Different from traditional approaches, RL control algorithm can dynamically adapt to the current traffic state according to the real-time road traffic environment. However, the traffic light control algorithm only based on the current traffic conditions may not be able to deal with the high complexity and high dynamics of the traffic system. Because it only assigns the green light source according to the current state, it ignores the congestion caused by the large traffic flow that may occur in the future for a period of time.

For the above reasons, we propose an algorithm named GPlight, which predicts the traffic flow in the future term before adopting the reinforcement learning algorithm to control the traffic lights. In GPlight algorithm, traffic forecast, which aims to estimate the urban traffic status of a period of time, is integrated with the traffic light control to reduce the possible future congestion, so as to improve control efficiency and avoid congestion. More specifically, in real-world intersection scenarios, each lane has a maximum vehicle capacity limit. Moreover, the green light duration is not infinite. During a limited period of green light,  some vehicles are bound to be left in the incoming lanes of the intersection. If congestion is expected in the future, traffic prediction can help the traffic light control system reduce the number of vehicles left behind in the congestion direction, freeing up more space for future vehicles heading into intersections.

Traffic prediction algorithms fall into two main categories. The first one is the traditional approaches based on statistics, and the other one is the deep learning algorithms driven by big data, such as Convolutional Neural Networks (CNN) for spatial correlation learning and Recurrent Neural Networks (RNN) for temporal sequence learning. However, the traditional convolution approaches cannot capture the structural information of the road network. On the other hand, the road network can be conveniently represented as a graph, which preserves its structural information. A state-of-the-art approach combines the network structure modeling using graph theory algorithm with the convolution algorithm, and proposes a Graph neural Network (GNN), which is effective in traffic state prediction\cite{6chengcheng2019dynamic,7zhiyong2019traffic,8shengnan2019attention,9kan2020optimized,10yuxuan2019gcgan,11zulong2019dynamic,12chen2019gated}.

In this paper, a new traffic light control algorithm named GPlight is proposed, which combines traffic prediction and reinforcement learning control. Our study considers the traffic light control problem of a complex traffic network with multiple intersections, aiming to increase the throughput of the road network. The proposed GPlight algorithm is divided into two stages. Firstly, the traffic flow information recorded in the previous period is used to predict the traffic flow in a short term. After that, the optimal green light direction and green light duration are selected for the intersection by comprehensively considering the traffic prediction results and the road traffic status obtained from real-time observation. To summarize, the main contributions of this work are as follows:

\begin{itemize}
\item We emphasize the importance of the traffic prediction in ITS and integrate it with traffic light control to reduce the traffic congestion.

\item We propose a GPlight algorithm, which the GNN algorithm is used to predict the traffic flow of road network in the future time, and an RL algorithm is used to control the traffic lights based on the predicted results. The duration of the green light is dynamically adjusted according to the predicted and current road congestion.

\item We conduct experiments on both synthetic and two real-world data-sets of Hangzhou and New-York. Extensive results demonstrate the effectiveness and rationality of the proposed algorithm.
\end{itemize}

%The remaining of the paper is as follows: First we review the related work in Section \ref{2} and then formalize the problem in Section \ref{3}. The GPlight algorithm is presented in Section \ref{4}. After conducting experiments in Section \ref{5}, we conclude the paper in Section \ref{6}.

\section{Related Work}\label{2}
There have been a lot of researches on intelligent traffic signal control problem using RL algorithm, which have achieved better performance than traditional approaches. Previous paper\cite{2zheng2019learning} presents an algorithm of traffic light phase control in single intersection environment. Some other papers\cite{3wei2019colight} consider the interaction and influence between adjacent intersections, extending the traffic light control system to the multi-intersection environment, and using a real-world road network model for experiments. A novel approach\cite{4wei2019presslight} has been introduced which combines RL algorithm with “max pressure” to get a more intuitive representation of the state and reward. However, all of these above algorithms rely only on the current state as the basis for action selection.  An approach that considers not only the current state but also the future when making decisions may yield a better strategy.

Some studies\cite{5kim2020cooperative} make traffic prediction of the next state according to the variables such as weather, date and time in the real world, and add the prediction results into the calculation of Q-function for traffic light control. However, this traditional approach basing on statistics cannot effectively reflect the urban traffic system with the characteristics of randomness and dynamic changes. Therefore, various deep learning algorithms, such as convolutional neural network (CNN) and recursive neural network (RNN), are increasingly used in traffic prediction problem, in which CNN is used to capture spatial dependency and RNN to temporal dynamic. However, these algorithms destroy the connectivity and structural relationship between the nodes in the complex traffic network and cannot capture the structural characteristic information in the network. State-of-the-art researches express the complex road network in the form of graph and combine it with neural network, proposing the graph neural network (GNN) algorithm\cite{6chengcheng2019dynamic,7zhiyong2019traffic,8shengnan2019attention,9kan2020optimized,10yuxuan2019gcgan,11zulong2019dynamic,12chen2019gated}. These studies use the directed or undirected graphs to define the nodes in the traffic network and the relations between them, constructing traffic prediction frameworks which use GNN to capture spatio-temporal characteristics of traffic flow data.

Because of its spatial characteristics, the traffic road network can be easily represented as a graph, in which a variety of node and edge definitions can be adopted. One way is to take each vehicle on the road as a node, and the connections between it and up to eight vehicles around as edges\cite{13frederik2019graph}. This model will cause high complexity in the road network with a large number of vehicles. Another approach is to use nodes to represent road sections, and adjacency matrix is used to represent whether roads are connected or not\cite{14xu2019road,15ling2019tgcn,16wang2018efficient}. A more easily implemented approach is to take sensors installed in the road network as nodes, and the edges and weights indicate the connectivity between sensors\cite{17ge2019temporal,18zhou2020reinforced,19bing2018spatio}. These approaches above are simple, but not suitable for traffic light control which get observations at intersections.

\section{Problem Definition}\label{3}
In this section, we will describe traffic flow prediction and traffic signal control problems, introduce the details of scenario modeling, and define some relevant terms and notations.

\subsection{Road Network Represented}\label{3.1}
We use a complex road network with multiple intersections as a scenario for traffic prediction and control. Multiple intersections of the road network use traffic lights to control the flow of traffic passing by, and every two intersections are connected by directed lanes, as shown in the Figure 1.

\begin{figure*}[t]
	\centering
	\includegraphics[width=0.6\linewidth]{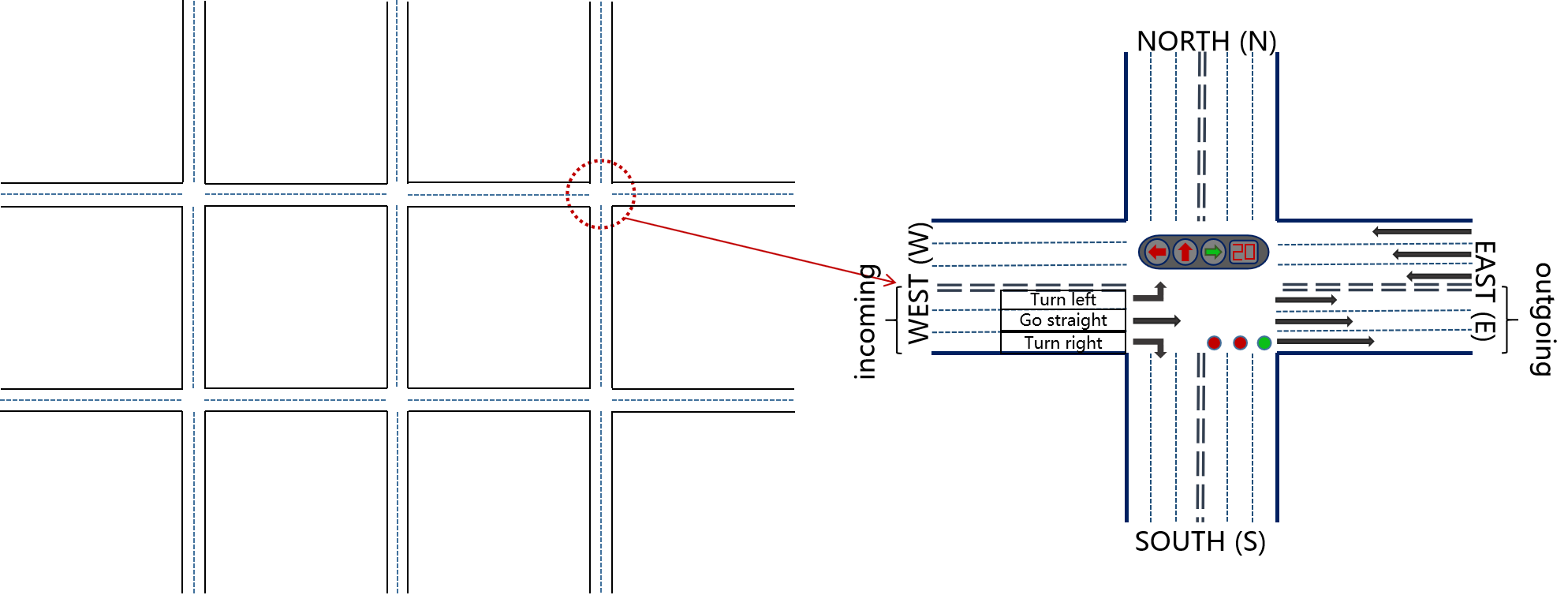}
	\caption{Road network structure and intersection setting}
	\label{intersection}
\end{figure*}

For each intersection, there are four approaches in four directions: east, west, north and south, denoted by “E”, “W”, “S”, “N”. This is one of the most common situations in real-world traffic. The approach in each direction is divided into two directions: incoming and outgoing. Vehicles approach the intersection from incoming lane, pass the intersection and then leave from outgoing lane. The incoming lane is also divided into three lanes: left, right and straight, which means the vehicles travelling on it will go in three different directions.

We set up an agent at each intersection, which uses three traffic lights (i.e. red, yellow and green) to control the traffic flow. We combine 8 kinds of traffic movement directions (that is, the vehicle passes through the intersection from the incoming lane to the outgoing lane) in pairs, and divide them into 4 phases. Each phase contains two non-conflicting traffic movement directions. At the green time, vehicles in the two directions corresponding to this phase are allowed to pass, while the remaining directions are set as red lights. Light the yellow light for 5 seconds after the green light duration to clear the vehicles passing through the intersection. The specific intersection Settings are shown in the Figure 1.

\subsection{Traffic Prediction Problem}\label{3.2}
We firstly describe traffic flow prediction and traffic signal control problems. For the traffic road network, we use a weighted undirected graph to represent it, which is noted as $\mathcal{G}=(V, E, W)$. $V=\{v_1, v_2, \cdots, v_N\}$ represents $N$ nodes as intersections in the road network, and the traffic feature data collected at the intersections is used for the training and testing of the prediction model. $E$ represents the edges in the undirected graph $\mathcal{G}$, represents the roads connecting the intersections, indicating the intersections' connectivity. For intersections $v_x$ and $v_y$, $e(x,y)$ has values $1$ and $0$. When $v_x$ and $v_y$ are connected, the value is $1$, otherwise $0$. And $W \in \mathbb{R}^{N \times N} $is the weighted adjacency matrix of graph $\mathcal{G}$. Specifically, the edge weight from $v_x$ to $v_y$ is noted as $w(x,y)$.

In the traffic road network, we regard each intersection as a node of the weighted undirected graph $\mathcal{G}$, and observe the traffic attribute characteristics of the road through sensors over a period of time. The observed data-set is expressed as $X \in \mathbb{R}^{T \times N \times D}$, where $T$ represents the sampling time, $N$ represents the number of nodes, and $D$ represents the dimension of observed traffic characteristics. Node attribute characteristics can be any traffic information, such as traffic flow, vehicle speed, etc. Specifically, we use $x_t^{i,d}$ to represent the information of the $d$-th feature in node $i$ observed at time $t$. Let the values of all the characteristics of node $i$ at time $t$ be expressed in terms of $x_t^i$, matrix $X_t=(x^1_t, x^2_t, \cdots, x^N_t ) \in \mathbb{R}^{N \times D}$ records all the characteristic information of all nodes in the graph at time $t$, and for a period of time, matrix $X=(X_{t-T+1}, X_{t-T+2}, \cdots, X_t) \in \mathbb{R}^{N \times D\times T}$is regarded as the feature matrix of the traffic network graph.

In this way, the graph-based traffic prediction problem is regarded as measuring the traffic feature information of $N$ nodes in the past $T$ time steps, and using the recorded observation information $X$ to predict the traffic features of nodes in the next $H$ time steps, which is denoted as $Y=(Y_{t+1}, Y_{t+2}, \cdots, Y_{t+H} ) \in \mathbb{R}^{N \times D \times H}$, as shown in Eq. 1:
\begin{equation}
\begin{split}
	\label{Eq1}
	(Y_{t+1}, Y_{t+2}, \cdots, Y_{t+H})=\arg\max P\{Y_{t+1}, Y_{t+2}, \\\cdots, Y_{t+H}\mid X_{t-T}, X_{t-T+1}, \cdots, X_t\}\ .
\end{split}
\end{equation}

\section{GPlight Algorithm}\label{4}
In this section, we first describe the algorithm of traffic prediction using GNN. Then, the framework of GPlight algorithm using RL algorithm to control the traffic flow at the intersection with the predicted results is presented.

\subsection{Traffic Prediction on Road Graph}\label{4.1}
The structure of the traffic prediction section at GPlight, as shown in Figure 2, consists of two spatial-temporal convolution blocks and a fully-connected layer, which are cascaded together. Each convolution block contains two gated temporal convolutional layers and a spatial graph convolutional layer in the middle of them. The spatio-temporal correlation information of traffic flow is extracted by convolution blocks, and the features obtained are integrated and processed by the fully-connected output layer to generate prediction.

\begin{figure*}[t]
	\centering
	\includegraphics[width=0.75\linewidth]{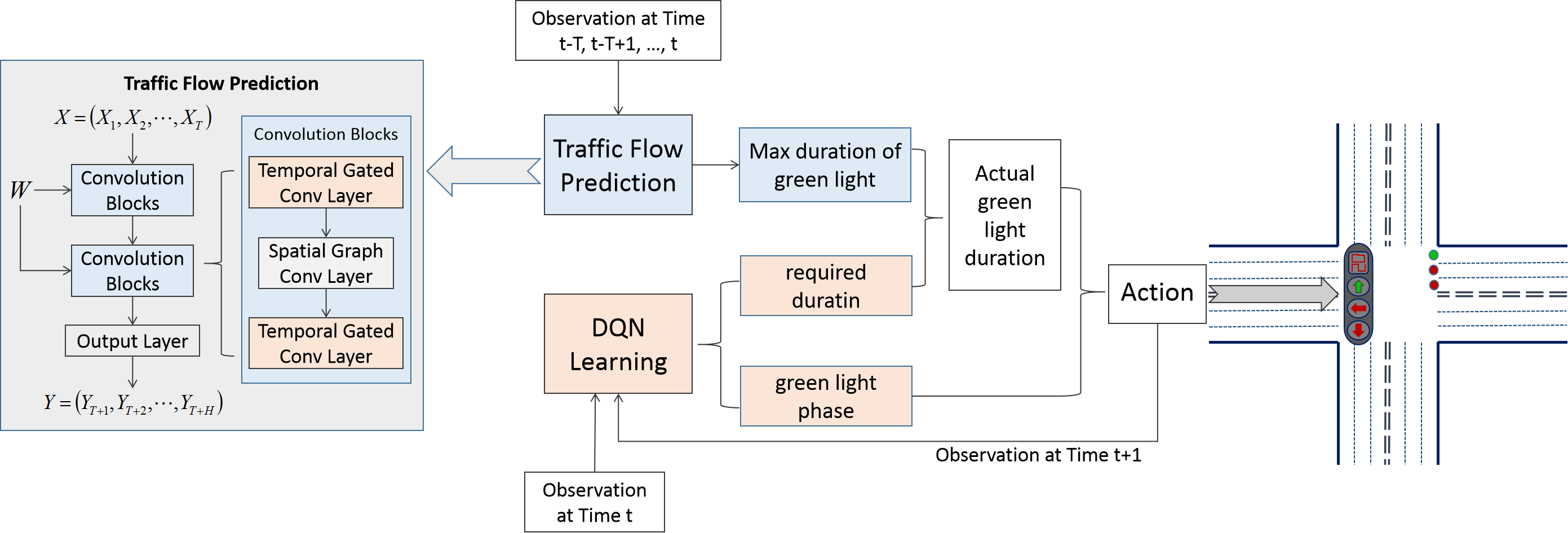}
	\caption{The framework of GPlight} \label{framework}
\end{figure*}

\paragraph{Spatial graph convolutional layer.} GPlight uses GNN to extract spatial information from previous traffic flow observations. The convolution is carried out directly on the data with graph structure, which preserves the spatial structure features of the traffic network. In order to apply the standard convolution to the graph structure, the Fourier transform is used to apply the convolution to the spectral domain, commonly known as the spectral graph convolution. In the concept of spectral convolution, the graph convolution operator is defined as follows, such as the convolution between the convolution kernel $\Theta$ and an input signal $x$:
\begin{equation}
\label{GCN}
\Theta \ast_{\mathcal{G}}x =U\Theta (\Lambda)U^Tx=\Theta (U\Lambda U^T)x=\Theta(L)x\ ,
\end{equation}
where $L=I_n-D^{-\frac{1}{2}}WD^{-\frac{1}{2}}\in\mathbb{R}^{N\times N}$ is the normalized Laplacian matrix, in which $I_n\in\mathbb{R}^{N \times N}$ is an identity matrix and $D\in\mathbb{R}^{N \times N}$ is the diagonal degree matrix with $D_{ii}=\sum_jW_{ij}$; $U\in\mathbb{R}^{N \times N}$ is the Fourier basis matrix of eigenvectors of $L$, and $\Lambda\in\mathbb{R}^{N \times N}$ is the diagonal matrix of the eigenvalues of $L$.

The convolution computation in large graph networks has high complexity, and Chebyshev polynomial approximation can be used to reduce the computational complexity of Eq. \ref{GCN}. In Chebyshev polynomial $T_k(x)$, the graph convolution kernel $\Theta$ can be approximately written as a polynomial: $\Theta(\Lambda)=\sum_{k=0}^{K-1}\theta_kT_k(\hat{\Lambda})$, where $\theta$ is the coefficient of the polynomial, and $K$ is the size of the convolution kernel. The diagonal matrix $\Lambda$ is scaled as $\hat{\Lambda}=2\Lambda/\lambda_{max}-I_n$, where $\lambda_{max}$ is the largest eigenvalue of $L$. In this way, the definition of convolution in Eq. 2 can be approximately written as:
\begin{equation}\label{Eq.3}
\Theta \ast_{\mathcal{G}}x=\sum_{k=0}^{K-1}\theta_kT_k(\hat{L})x\ ,
\end{equation}
where $\hat{L}=2L/\lambda_{max}-I_n$ denotes the scaled Laplace matrix.

\paragraph{Temporal gated convolutional layer.} The temporal convolutional layer is used to capture the temporal characteristics of the traffic flow, which consists of a one-dimensional causal convolution and a gated unit. Let $Y\in\mathbb{R}^{m\times c_{in}}$ represents the input to the time convolutional layer, where $m$ represents the size of temporal, and $c_{in}$ represents channel dimensions. The temporal gated convolution can be defined as:
\begin{equation}\label{Eq.4}
\Gamma\ast_{\tau}Y=Y_1\odot\sigma (Y_2)\ ,
\end{equation}
where $\tau\in\mathbb{R}^{K\times c_{in}\times2c_{out}}$ is the width-$K$ convolutional kernel, and $Y_1, Y_2\in\mathbb{R}^{(m-K+1)\times c_{out}}$ are the input of GLU. $\odot$ is the element-wise Hadamard product, and $\sigma(Q)$ is the sigmoid gate which controls input $Y_1$ of the current states.

\subsection{DQN Algrithom Setting}\label{4.2}
Deep Q-Network (DQN) is used for traffic light control, which combines Q-learning with deep neural network. In our scene, an DQN agent is set at each intersection in the road network, and trains its own model separately. Considering that the traffic flow between intersections and the action selections of traffic lights will influence each other, and closer intersections have a bigger impact, the agents obtain the information of the adjacent intersections through the attention mechanism, and use these information to realize the collaborative control of multiple intersections.

A Markov Decision Process (MDP) is used to represent the process of agents making decisions in its interaction with the environment. This MDP can be represented by $<S, A, p, R, \gamma>$, in which there is a set of states $s_t\in S$, a set of actions $a_t\in A$, a transfer probability $p$, a reward function $R$, and a discount factor $\gamma$.

In GPlight algorithm, the state represents the observation of the situation at intersections. The action space includes the choice of both green phase direction and green phase duration. Inspired by the pressure algorithm in Presslight\cite{4wei2019presslight}, this paper uses an improved pressure algorithm to define the rewards for actions. The improved pressure algorithm takes into account the maximum carrying capacity of the lane, and is determined by the number of vehicles on the incoming and outgoing lanes, which is calculated by the following equation:
\begin{equation}
\label{P_i}
P_i = N_{in}*(1- \frac{N_{out}}{N_{max}})\ ,
\end{equation}
where $P_i$ is the pressure of traffic movement $i$. $N_{in}$ and $N_{out}$ is the number of vehicles on incoming and outgoing lane, respectively. $N_{max}$ is the maximum number of vehicles that can fit in a lane.

The reward for traffic movement $i$ is set according to the pressure in Eq. 5, which is defined as:
\begin{equation}
\label{6}
r_i=-P_i\ .
\end{equation}
Since there are 12 traffic movements at one intersection, the total reward $R$ for an action is:
\begin{equation}
\label{7}
R=\sum_{i=0}^{11}{r_i}=-\sum_{i=0}^{11}{P_i}\ .
\end{equation}

At each intersection, the agent takes the phase information and vehicle queue length on incoming and outgoing lanes at time $t$ as the current state $s_t$, estimates Q-value of all actions according to $s_t$, and selects the green light phase and duration as action $a_t$. Then the agent observes the state $s_{t+1}$ at time $t+1$ and gets feedback to update the parameters of Q-network by gradient descent. The Q-function is defined as:
\begin{equation}\label{8}
Q(s_t, a_t)=R(s_t, a_t)+\gamma*\max_{a_{t+1}}\{Q(s_{t+1}, a_{t+1})\}\ ,
\end{equation}
where $\gamma$ is the discount factor. And the loss function is expressed as follow:
\begin{equation}
\label{9}
\begin{split}
J=\sum \frac{1}{B}
(R_t+\gamma \max\hat{Q}(s_{t+1},a_{t+1};\hat{\theta})-Q(s_t,a_t;\theta))^2\ ,
\end{split}
\end{equation}
where $\hat{Q}$ is the target function, and $Q$ is the primary function. $B$ is the batch size in DQN.

\subsection{Framework of GPlight}\label{4.3}
The GPlight framework is divided into two parts: traffic flow prediction and traffic light control using RL algorithm, as shown in Figure 2. In GPlight, GNN algorithm introduced in Section 4.1 is used to predict the traffic flow in a short term. This prediction is used to determine the expected green light duration $t_{exp}$ at the intersection. Specifically, the number of vehicles coming in the future is used to determine the duration of the green light that will allow all future vehicles in chosen direction to pass. Then, DQN agent chooses the required duration of green light $t_{req}$ at intersection based on the road surface observation information at current time, which represents the duration of green time required to allow all vehicles currently waiting on the incoming lane to pass. Both the values of $t_{exp}$ and $t_{req}$ can be calculated by the number of vehicles, the average speed and acceleration of vehicles, and the length of intersections. The actual green light duration is determined by the minimum of $t_{exp}$ and $t_{req}$. In this way, the agents are able to determine the current green light duration based on the number of vehicles that will arrive in the future, choose a larger green light duration when the congestion is coming, so that the traffic flow in this direction can be cleared in advance and more space for the coming vehicles will be made. On the other hand, the choice of the minimum value in $t_{exp}$ and $t_{req}$ also avoids the waste of green light resources caused by too long duration. Finally, the agents select the optimal green phase and duration as the action based on the current observation and the prediction of the future traffic flow, and executes it in the next time step.

The pseudocode of GPlight algorithm is shown in Algorithm 1.

\begin{algorithm}[t]
	\caption{GPlight: Traffic Light Control based on GNN Prediction}
	\label{alg:GPlight}
	\begin{algorithmic}
		\Require
		Graph $\mathcal{G}=(V, E, W)$, episode length $T$, greedy $\epsilon$, update rate $\alpha$, target network replacement frequency $C$
		\State Initialize $Q$ with parameters $\theta$, $\hat{Q}$ with parameters $\hat{\theta}$
		\For{each $episode$} 
		\State Initialize step number $t$ and total time $t_{sum}$ to be $0$
		\While{$t_{sum} < T$}
		\State /*\ \ \ Traffic prediction\ \ \ */
		\State Capture spatial features by Eq. \ref{Eq.3}
		\State Capture temporal features by Eq. \ref{Eq.4}
		\State Get prediction: $(Y_{t+1}, Y_{t+2}, \cdots, Y_{t+H})\gets(X_{t-T}, X_{t-T+1}, \cdots, X_t)$
		\State Get expected green phase duration $t_{exp}\gets(Y_{t+1}, Y_{t+2}, \cdots, Y_{t+H})$	
		\State /*\ \ \ Traffic light control\ \ \ */	
		\State Select a random phase  $pha$ with probability $\epsilon$
		\State Otherwise $pha \gets \mathop{\arg\max}_{pha}{Q(s_t,pha;\theta)}$
		\State Get required green phase duration $t_{req}$ by current observation  
		\State $t_{green}=\min\left(t_{exp}, t_{req}\right)$
		\State $a_t \gets \{pha, t_{green}\}$
		\State Execute $a_t$, observe new state $s_{t+1}$, get reward $R$
		\State $t_{sum} \gets t_{sum}+t_{green},\ t \gets t+1$
		\State Calculate the loss $J$ by Eq. \ref{9}
		\State Update $\theta$ with $\nabla J$
		\State Every $C$ steps update $\hat{Q}$:\quad $\hat{Q} \gets Q$
		\EndWhile
		\EndFor
	\end{algorithmic}
\end{algorithm}

\begin{table*}[htb]
	\centering
	\begin{tabular}{l|ccc||ccc}
		\toprule
		& \multicolumn{3}{c||}{Average Travel Time} & \multicolumn{3}{c}{Throughput} 		\\
		\hline
		& Single & Hangzhou &  New-York	& Single & Hangzhou &  New-York   \\
		\hline
		FixedTime   & 135.79 & 249.72  & 120.94 
		            & 1250   & 3410    & 200   \\ \hline
		MaxPressure & 212.66 & 346.33  & 397.62
		            & 2082   & 4394    & 2373 \\
		\hline
		CoLight-Fixed    & 151.01 &  365.54  & 187.20
		                 & 2178 &  4473    & 2716 \\
		\hline
		CoLight-Dynamic    & 130.38 & 355.87 & 183.72
		                   & 2076 & 4383   & 2713  \\
		\hline
		PressLight-Fixed   & 105.46 & 374.62 & 454.90  
		                   & 2210  & 4359   & 1312  \\
		\hline
		PressLight-Dynamic & 124.04  &  381.28 & 363.04
		                   & 2096 &  4397   & 1206 \\
		\hline
		GPlight            & \textbf{91.54} &\textbf{336.78} &  \textbf{181.81}
		                   & \textbf{2288}  & \textbf{4575}   &  \textbf{2718}   \\
		\bottomrule
	\end{tabular}
	\caption{Average Travel Time and Throughput in one hour}
	\label{tab_time_maxnum}
\end{table*}

\iffalse
\begin{table*}[htb]
	\centering
	\begin{tabular}{l|ccc||ccc}
		\toprule
		& \multicolumn{3}{c||}{Average Travel Time} & \multicolumn{3}{c}{Throughput} 		\\
		\hline
		& Single & Hangzhou &  New-York	& Single & Hangzhou &  New-York   \\
		\hline
		FixedTime   & 460.94 & 557.68  & 1846.21
		& 1250   & 3410    & 200   \\ \hline
		MaxPressure & \textbf{211.36} & 406.35  & 421.27
		& 2082   & 4394    & 2373 \\
		\hline
		CoLight-Fixed    & &  358.71  & \textbf{177.50}
		& &  4360    & 2699 \\
		\hline
		CoLight-Dynamic    & & 417.77 & 177.95
		& & 4183   & 2712  \\
		\hline
		PressLight-Fixed   & & 390.42 & 848.51
		& & 4359   & 1243  \\
		\hline
		PressLight-Dynamic & &  380.69 & 897.26
		& &  4348   & 1206 \\
		\hline
		GPlight            & 238.49 & \textbf{342.00} &  179.06
		& \textbf{2288}  & \textbf{4575}   &  \textbf{2718}   \\
		\bottomrule
	\end{tabular}
	\caption{Average Travel Time and Throughput in one hour}
	\label{tab_time_maxnum_}
\end{table*}

\fi
\begin{figure*}[htb]
	\centering
	\subfigure[Single]{\includegraphics[width=0.2\linewidth]{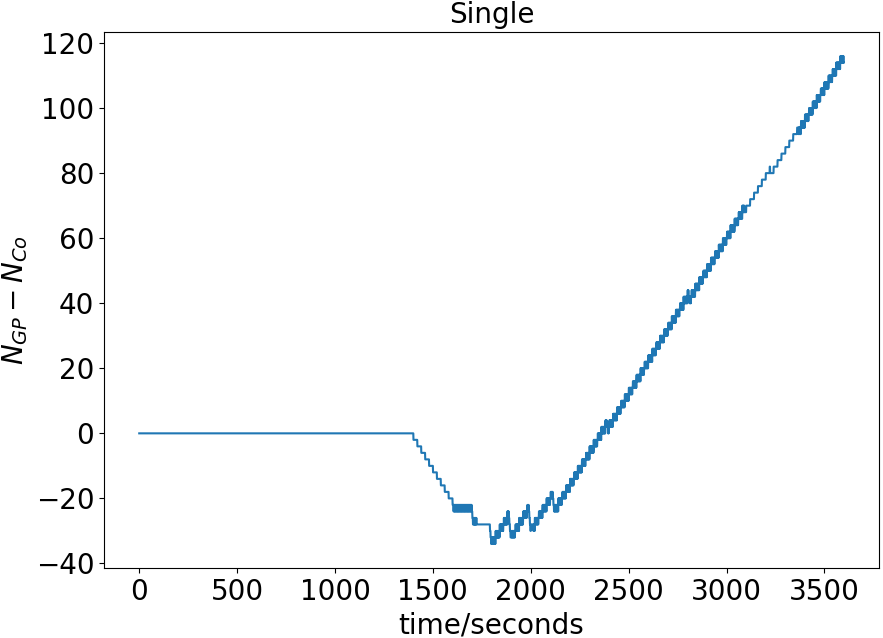}}
	\subfigure[New-York]{\includegraphics[width=0.2\linewidth]{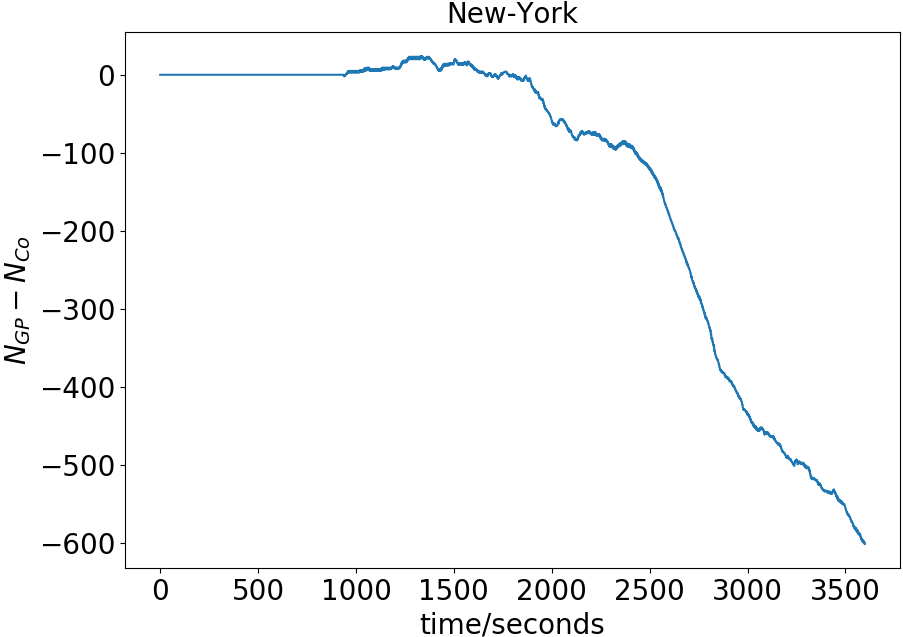}}
	\subfigure[Hangzhou]{\includegraphics[width=0.2\linewidth]{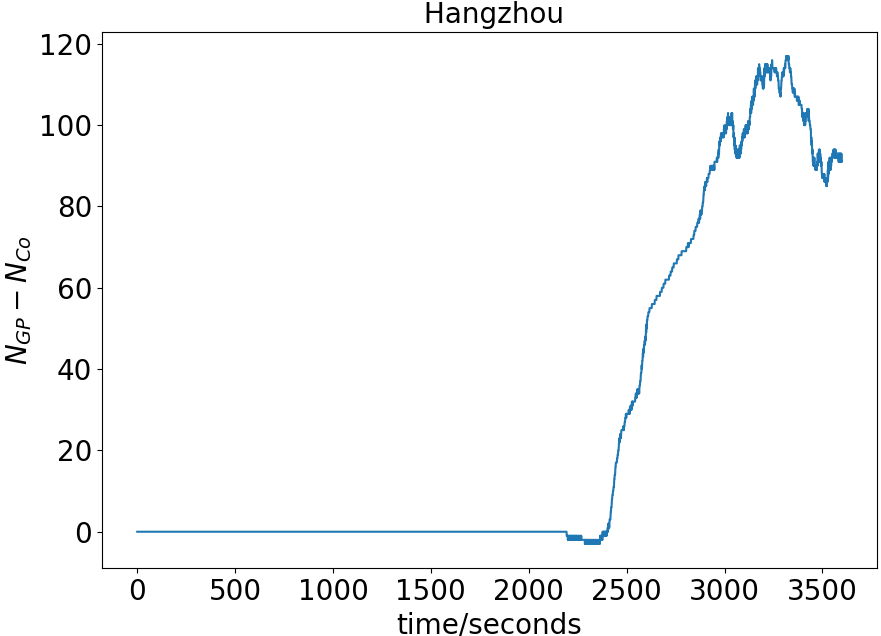}}
	\caption{Gap of number of passed vehicles in one hour }
	\label{fig_count}
\end{figure*}

\section{Experiment and Analysis}\label{5}
In this section, based on CityFlow\cite{20ZhangCityFlow} simulator, the simulation results are shown to verify the effectiveness of the proposed GPlight algorithm, which is compared with several state-of-the-art algorithms.

\subsection{Data-sets and Experiment Setting}\label{5.1}
Three data-sets are used in traffic light control: Single, New-York and Hangzhou. For the Single data-set, there is only one intersection and the traffic flow is generated manually. Except for the flow going straight from west to east and from east to west, all other have an  interval of 20 seconds. The interval of the W-E and E-W traffic flow is 1 second from 900-th second to 2700-th second and 20 seconds in other time. New-York and Hangzhou data-sets are based on collected vehicle trajectories from practice. There are 48 intersections in New-York, the lane length for the WE and NS direction is $350m$ and $100m$ respectively; and there are 16 intersections in Hangzhou, the distance between adjacent intersections is set at $300m$, which are the same setup as in Colight\cite{3wei2019colight}.

The GNN module for traffic flow prediction is trained by using traffic flow data completely consistent with the topology of traffic light control scene, and then is integrated into the traffic light control part. The distance between each node is calculated according to the respective road network. The data of past 10 minutes is utilized to predict the traffic volume of next 5 minutes. For the history data, the maximum number of vehicles of each lane is calculated for every minute and is then input to the GNN module. The output of GNN is regarded as the maximum number of vehicles of next 5 minutes and is used to get the maximum duration of the green light.

In the experiment, the greedy $\epsilon$ in Algorithm 1 is decreasing from $0.8$ to $0.2$. The discount factor $\gamma$ for calculating the accumulated reward is set as $0.8$. The learning rate of the Q-network is set as $0.001$.

\subsection{Baseline}
\begin{itemize}
	\item \textbf{FixedTime}. Set green light for all phases with a pre-determined order.
	\item \textbf{MaxPressure} \cite{Varaiya2013MaxPressure}. Set green light for the phase with the max pressure
	\item \textbf{CoLight} \cite{3wei2019colight}.  An RL traffic light control algorithm for large-scale road networks. We consider two traffic light settings: Fixed and Dynamic. In Fixed, the traffic light duration is constant while in Dynamic the traffic light duration changes according to the real time traffic conditions.
	\item \textbf{PressLight} \cite{4wei2019presslight}. An RL traffic light control algorithm with pressure as the reward. Both Fixed and Dynamic are performed.
\end{itemize}

\subsection{Result and Analysis}\label{5.2}
In Table 1, we list both the average travel time and the throughput for the three data-sets in one hour. As has been stated, with the prediction of future traffic flow, GPlight has the potential to increase the throughput of the traffic network and decrease the travel time  by adjusting the maximum of green light duration. It seems that the FixedTime algorithm achieves the lowest travel time in table \ref{tab_time_maxnum}. However, this low travel time comes at the expense of low capacity, which means the number of vehicles passing the road network is too low to be a satisfactory solution.

%On the New-York data-set, however, GPlight achieves less throughput than Colight. This can be explained by the road network structure of the New-York data-set. The road is grid-shape with 16 intersections on the N-S direction and 3 on the W-E direction. The distance between each two adjecent intersections are 100 and 350 meters for the N-S and W-E respectively. The Hangzhou data-set, on the other hand, has both 4 intersections on W-E and N-S directions with distance being 800 and 600 meters respectively. The most distinctive difference between these two data-sets is thus the network structure. Due to the short lanes in New York, the maximum carrying capacity of a lane is also small and there will be less space reserved for future congestion by means of advance release. In contrast, longer lanes such as those in Hangzhou, will be able to carry more vehicles, freeing up more space for future congestion through early release. In other words, the effect of prediction can be better reflected on longer lanes. In addition, the asymmetric and short-distance characteristic of New-York puts bigger challenge for the coordination of all intersections. An inaccurate prediction or a long green light time  will have a broader impact for the whole network. Since the lane between two adjecent intersections is short, a long green light time together with large traffic volume will cause congestion on not only this lane but also  its followed lanes. As a result, GPlight on New-York data-set fails to compete Colight.

To present a more clear compare, Figure 3 shows the gap  between GPlight and Colight over the process. $N_{GP}$ and $N_{Co}$ represent the number of vehicles passed using GPlight and Colight respectively. It can be seen that the two algorithms show no difference at first when the traffic is not heavy. As the manual added vehicles drive into the network from 900-th seconds, GPlight displays an edge. The same performance at the beginning and the followed positive gap demonstrates that GPlight works as well as Colight when the traffic is not heavy and increases the throughput when the traffic is heavy. It is also worth noting that $N_{GP} - N_{Co}$ first changes to negative and then positive. This will be analyzed later in the case study.

\begin{figure*}[htb]
	\centering
	\subfigure[Single]{\includegraphics[width=0.2\linewidth]{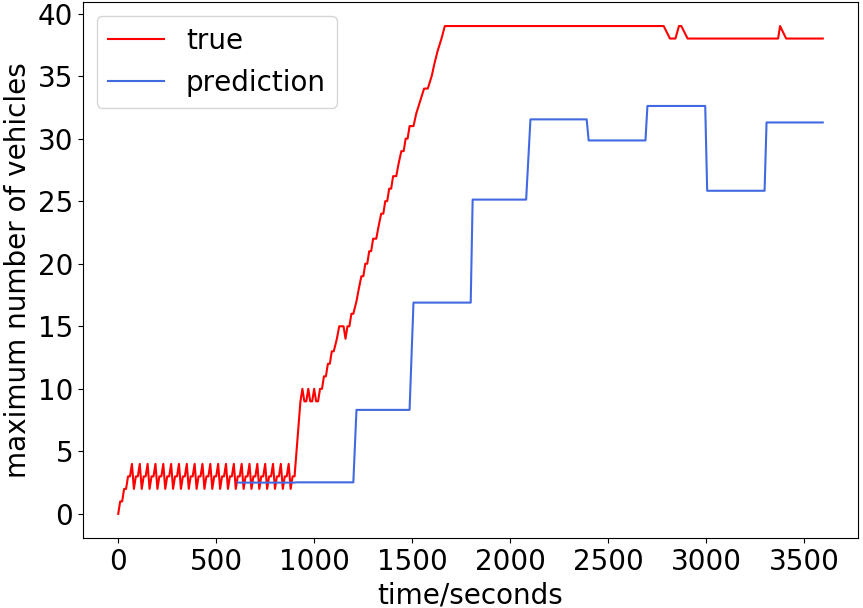}}
	\subfigure[newyork-inter-4]{\includegraphics[width=0.2\linewidth]{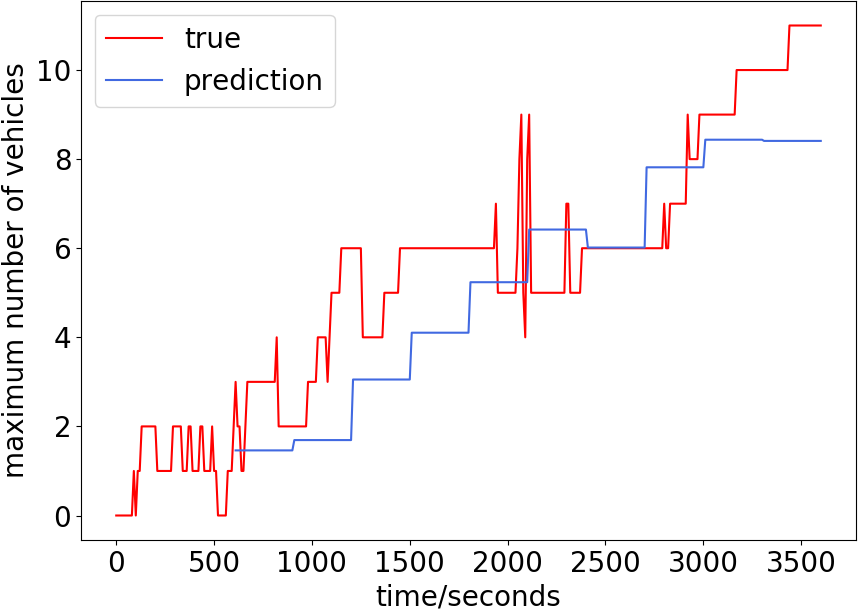}}
	\subfigure[newyork-inter-11]{\includegraphics[width=0.2\linewidth]{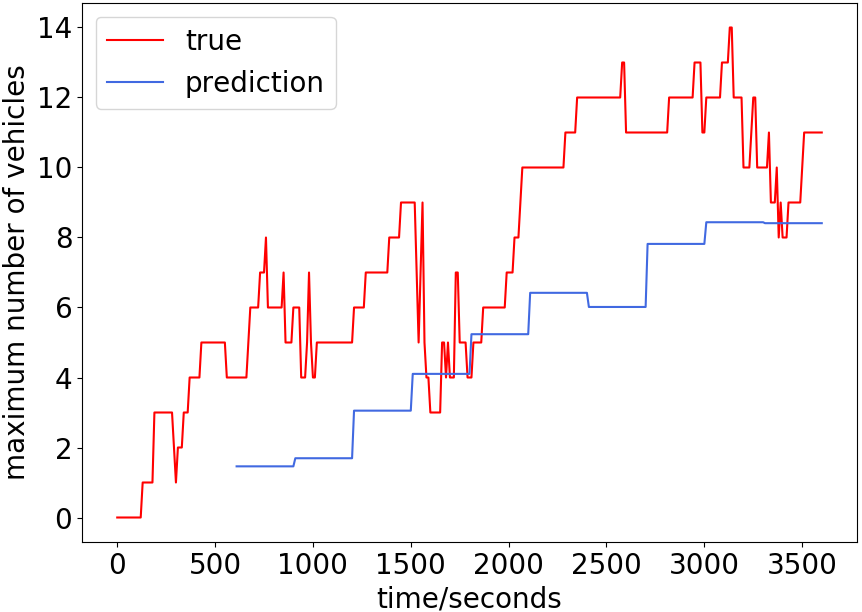}}
	
	\subfigure[Hangzhou-inter-1]{\includegraphics[width=0.2\linewidth]{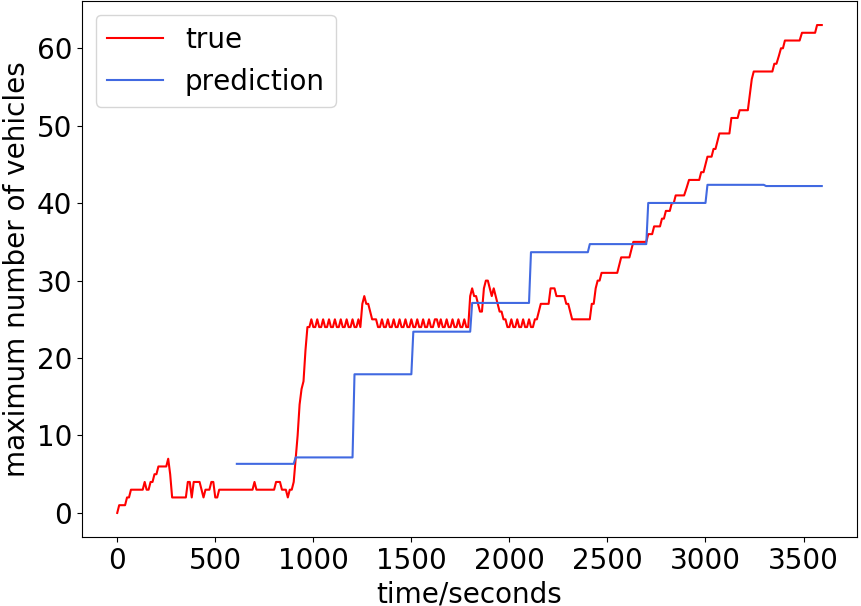}}
	\subfigure[Hangzhou-inter-6]{\includegraphics[width=0.2\linewidth]{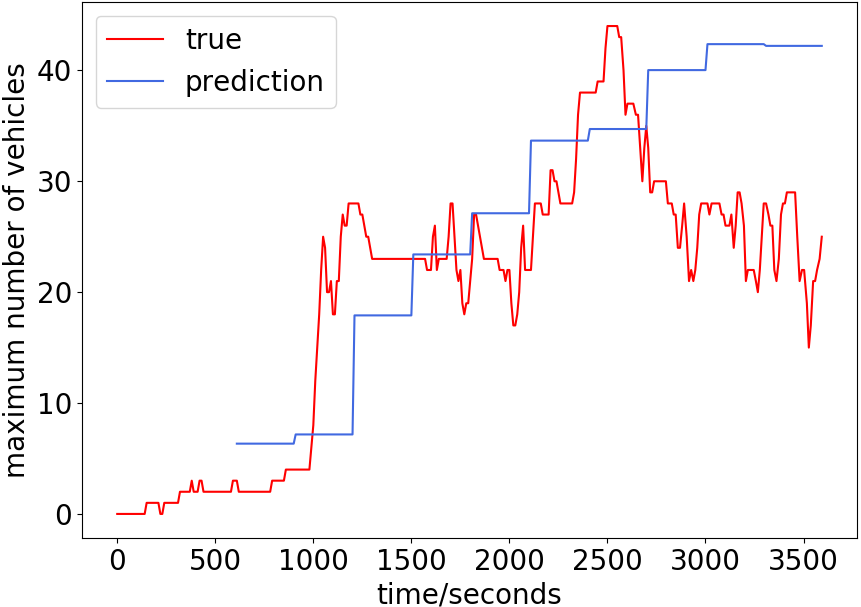}}
	\subfigure[Hangzhou-inter-10]{\includegraphics[width=0.2\linewidth]{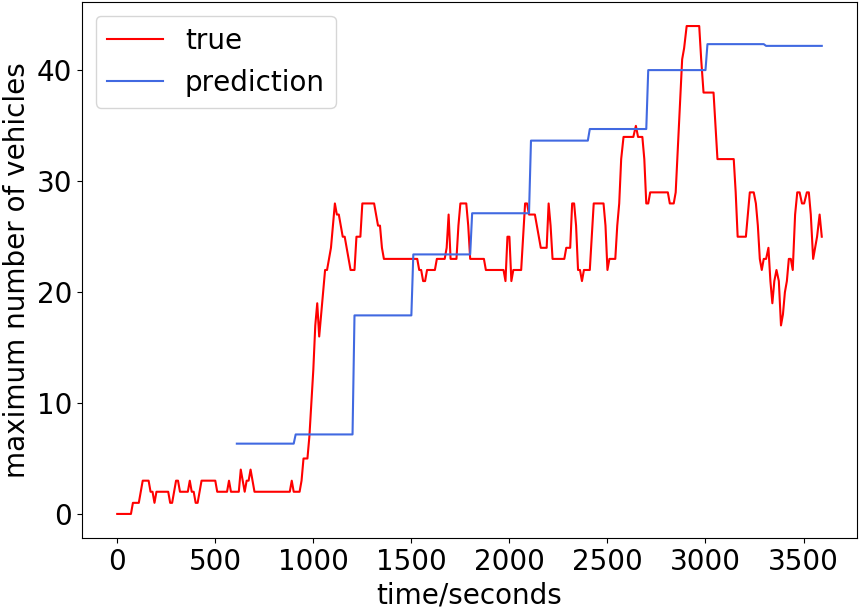}}
	\caption{Predicted and real traffic volume}
	\label{fig_pred_true-hangzhou}
\end{figure*}

\begin{figure*}[htb]
	\centering
	\subfigure[Single]{\includegraphics[width=0.2\linewidth]{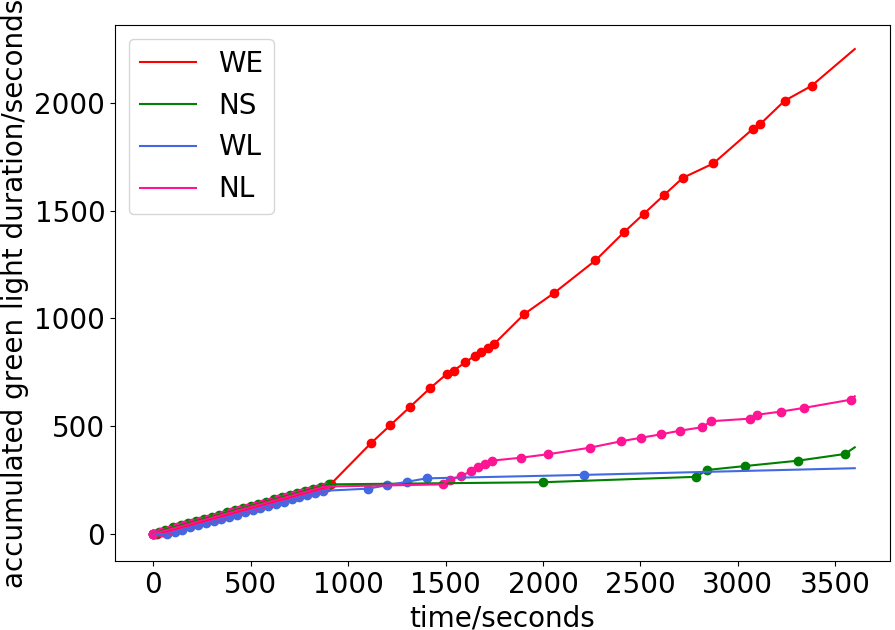}}
	\subfigure[New-York-inter-4]{\includegraphics[width=0.2\linewidth]{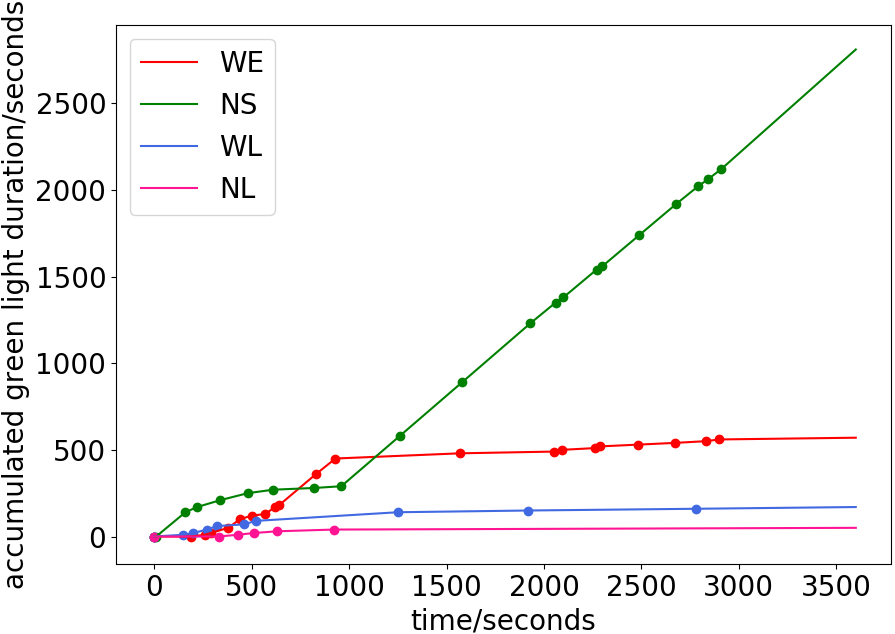}}
	\subfigure[New-York-inter-11]{\includegraphics[width=0.2\linewidth]{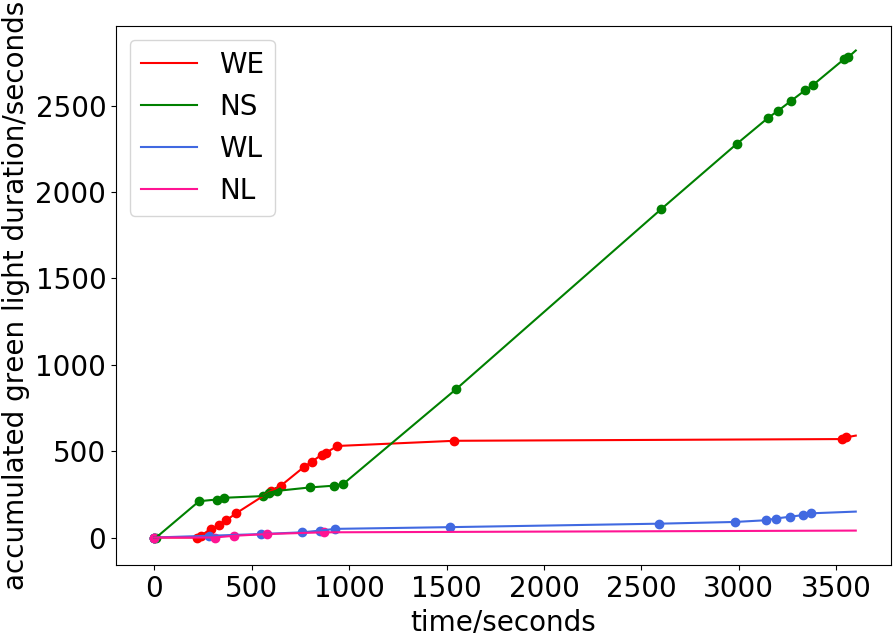}}
	
	\subfigure[Hangzhou-inter-1]{\includegraphics[width=0.2\linewidth]{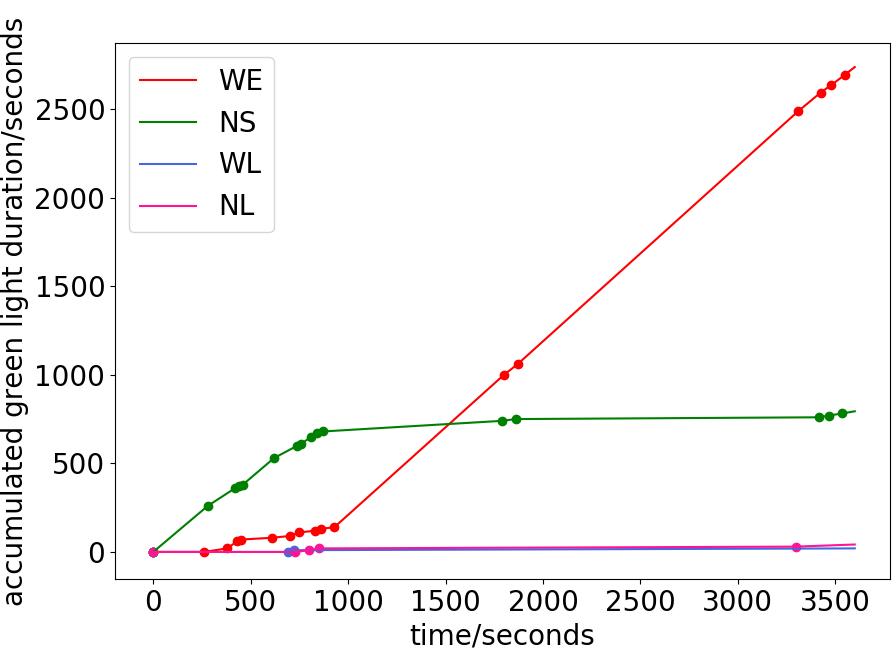}}
	\subfigure[Hangzhou-inter-6]{\includegraphics[width=0.2\linewidth]{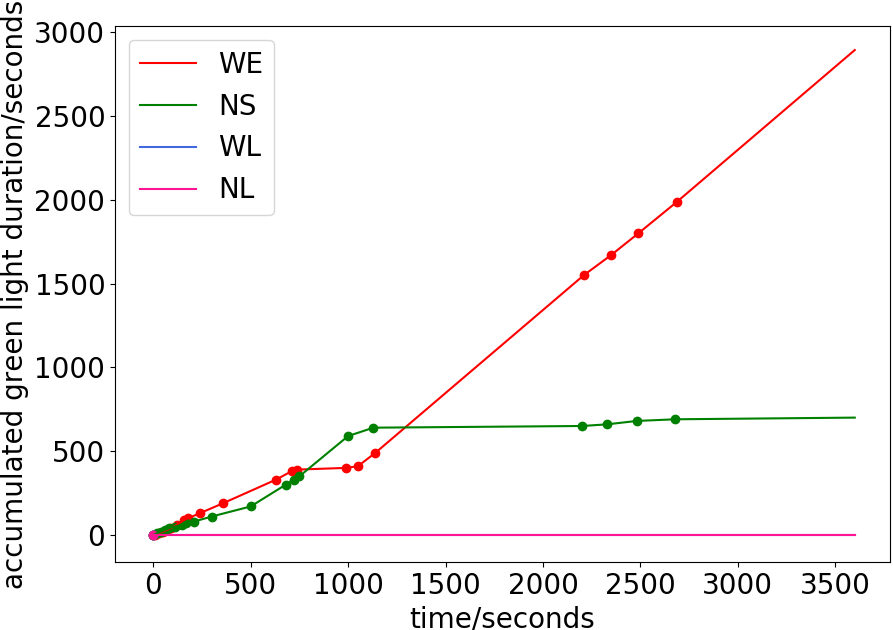}}
	\subfigure[Hangzhou-inter-10]{\includegraphics[width=0.2\linewidth]{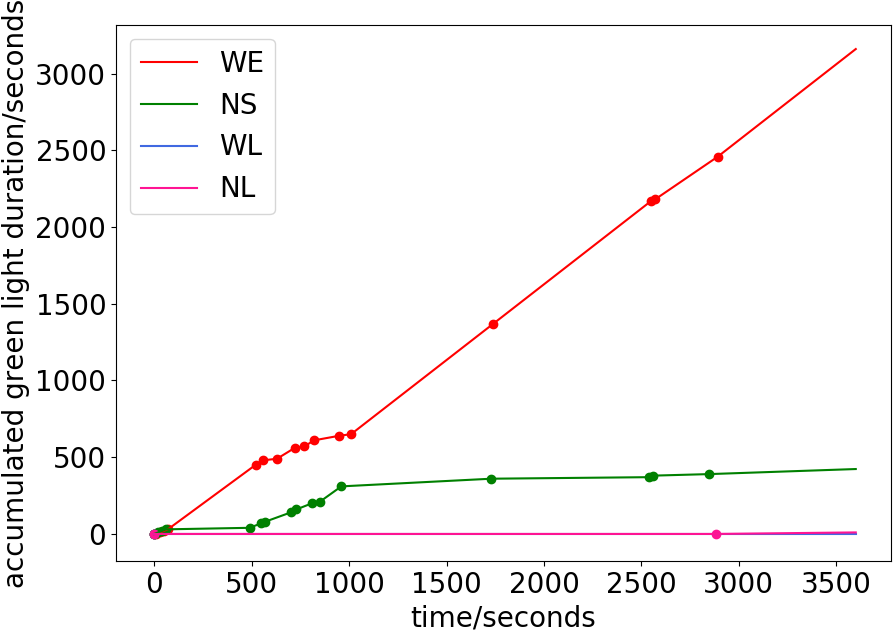}}
	\caption{Accumulated duration of four green light phase}
	\label{fig_lighttime-hangzhou}
\end{figure*}

\subsection{Case Study}\label{5.3}
To gain a more clear and intuitive understanding of how the GPlight makes choices and controls traffic flow, a manual vehicle flow is added from 900-th seconds to 2700-th seconds with the interval being 1 second to model heavy traffic in all data-sets in this subsection.

\textbf{Prediction of traffic volume}. An accurate prediction of the future traffic volume is important, because the maximum green light duration will adjust according to the predicted number of vehicles. Figure 4 draws the \textit{predicted} and \textit{real} maximum number of vehicles on three data-sets. The \textit{prediction} curve begins at around 600-th seconds since there is not enough history data to be input to the GNN during the first 10 minutes. The \textit{prediction} and \textit{real} curves show approximate trend with the \textit{real} curve being more variant. 

One can observe a delay from the \textit{real} and \textit{prediction}: The \textit{real} surges at around 1000-th seconds as the manual flow drives to the network, while the \textit{prediction} increases at around 1200-th seconds. The main reason can be summarized as follows. The prediction is based on the history data, and if the history shows no clear tendency, it will be unreasonable to predict a heavy traffic. Before the arrival of manual vehicle flow, the origin data-set presents little sign of heavy traffic. The manual flow arrives at 900-th second and such traffic feature is captured and input to the GNN after 5 minutes(300 seconds) according to the experiment setting. The prediction thus increases at 1200-th second. In practice, however, data of last day or last week can help to predict the traffic peak and avoid such delay. 

\textbf{Choice and duration of green light}. The GPlight controls the traffic flow by deciding the phase and duration of the green light. A detailed survey into the choice of GPlight can be therefore helpful and instructive. Figure 5 gives the accumulated duration of four phases over time.  Each dot on the curves represents a change of choice to its corresponding phase. Note that the algorithm maybe choose one phase repetitively and such repetitive choice is not dotted to present a more clear result. `WE' and `NS' represent the green light is set for the west-east and north-south straight direction respectively. `WL' means turning left from west to north and from east to south is allowed. `NL' means turning left from north to east and from south to west is allowed. From all the figures, it can be seen that after the manual vehicle flow drives into the traffic network, green light for the ``WE'' direction in Single and Hangzhou data-sets and for the ``NS'' direction in New-York data-set is set more frequently and longer.

Recall that the curve in Figure 3 first drops before rises. The choice of green light in Figure 5 can serve as a reasonable explanation. The expected rise can be explained by the long duration of the `WE' green light. Since the heavy manual flow is added to the west-east straight direction, the green light in this direction can encourage more vehicles to pass. As for the drop of the curve in Figure 3, take a look at the choice of green lights around the 900-th second. For the Hangzhou data-set, before the arrival of heavy traffic, more green lights are given to the other three directions, especially the ``NS'' one. Considering that a large number of vehicles will arrive on the ``WE'' direction and the green light will set to that direction for a relatively long period, it seams wise to set green lights for other directions to clear vehicles on those directions. Otherwise, vehicles that have already been on the lane of the other three directions would have to wait for a long time, which is unsatisfactory in practice. On the other hand, for the Single data-set, during the first 900 seconds, the four curves are approximately the same, which is reasonable as the intervals of the flow in the four directions are the same. Before the manual flow arrives, no preference is given to the other three directions and after the manual arrives, ``WE'' phase gets selected more frequently. As a result, the negative gap in Figure 3(a) is larger.

\section{Conclusion}\label{6}
In this paper, we propose GPlight, a reinforcement learning algorithm that combines traffic prediction and traffic light control for intelligent traffic control problem. The algorithm proposed first uses GNN, which combines graph theory with convolutional neural network algorithm, to predict the traffic flow in the short-term in future. After that, the prediction information and the real-time observation are used in the traffic light control using DQN algorithm. Experiments on both simulated and real-world data-sets show that the proposed GPlight algorithm improves the throughput and delay of the traffic network compared to the baseline algorithm.

\section*{Broader Impact}
This paper combines traffic prediction with intelligent traffic light control. The proposed algorithm uses the results of traffic prediction to clear the road which is going to become congested in advance. This algorithm can alleviate traffic congestion, reduce vehicle travel time, and thus reduce energy consumption and pollution.

\bibliography{mycite}

\begin{thebibliography}{21}
\providecommand{\natexlab}[1]{#1}
\providecommand{\url}[1]{\texttt{#1}}
\providecommand{\urlprefix}{URL }
\expandafter\ifx\csname urlstyle\endcsname\relax
  \providecommand{\doi}[1]{doi:\discretionary{}{}{}#1}\else
  \providecommand{\doi}{doi:\discretionary{}{}{}\begingroup
  \urlstyle{rm}\Url}\fi

\bibitem[{Chen et~al.(2019)Chen, Li, Teo, Zou, Wang, Wang, and
  Zeng}]{12chen2019gated}
Chen, C.; Li, K.; Teo, S.~G.; Zou, X.; Wang, K.; Wang, J.; and Zeng, Z. 2019.
\newblock Gated Residual Recurrent Graph Neural Networks for Traffic
  Prediction.
\newblock In \emph{Proceedings of the Thirty-Third AAAI Conference on
  Artificial Intelligence (AAAI-19)}. Hawaii, USA: AAAI.

\bibitem[{Chengcheng, Bo, and Xiaoping(2019)}]{6chengcheng2019dynamic}
Chengcheng, J.; Bo, W.; and Xiaoping, Z. 2019.
\newblock Dynamic Spatiotemporal Graph Neural Network with Tensor Network.
\newblock ArXiv preprint arXiv:2003.08729.

\bibitem[{Diao et~al.(2019)Diao, Wang, Zhang, Liu, Xie, and
  He}]{11zulong2019dynamic}
Diao, Z.; Wang, X.; Zhang, D.; Liu, Y.; Xie, K.; and He, S. 2019.
\newblock Dynamic Spatial-Temporal Graph Convolutional Neural Networks for
  Traffic Forecasting.
\newblock In \emph{Proceedings of the Thirty-Third AAAI Conference on
  Artificial Intelligence (AAAI-19)}. Hawaii, USA: AAAI.

\bibitem[{Diehl et~al.(2019)Diehl, Brunner, Le, and
  Knoll}]{13frederik2019graph}
Diehl, F.; Brunner, T.; Le, M.~T.; and Knoll, A. 2019.
\newblock Graph Neural Networks for Modelling Traffic Participant Interaction.
\newblock ArXiv preprint arXiv:1903.01254.

\bibitem[{Ge et~al.(2019)Ge, Li, Liu, and Zhou}]{17ge2019temporal}
Ge, L.; Li, H.; Liu, J.; and Zhou, A. 2019.
\newblock Temporal Graph Convolutional Networks for Traffic Speed Prediction
  Considering External Factors.
\newblock In \emph{Proceedings of the 20th IEEE International Conference on
  Mobile Data Management (MDM)}.

\bibitem[{Guo et~al.(2020)Guo, Hu, Qian, Liu, Zhang, Sun, Gao, and
  Yin}]{9kan2020optimized}
Guo, K.; Hu, Y.; Qian, Z.; Liu, H.; Zhang, K.; Sun, Y.; Gao, J.; and Yin, B.
  2020.
\newblock Optimized Graph Convolution Recurrent Neural Network for Traffic
  Prediction.
\newblock \emph{IEEE Transactions on Intelligent Transportation Systems} 1--12.
\newblock Early Access Article.

\bibitem[{Kim and Jeong(2020)}]{5kim2020cooperative}
Kim, D.; and Jeong, O. 2020.
\newblock Cooperative traffic signal control with traffic flow prediction in
  multi-intersection.
\newblock \emph{Sensors} 20(1): 137.

\bibitem[{Shengnan et~al.(2019)Shengnan, Youfang, Ning, Chao, and
  Huaiyu}]{8shengnan2019attention}
Shengnan, G.; Youfang, L.; Ning, F.; Chao, S.; and Huaiyu, W. 2019.
\newblock Attention Based Spatial-Temporal Graph Convolutional Networks for
  Traffic Flow Forecasting.
\newblock In \emph{Proceedings of the Thirty-Third AAAI Conference on
  Artificial Intelligence (AAAI-19)}. Hawaii, USA: AAAI.

\bibitem[{Varaiya(2013)}]{Varaiya2013MaxPressure}
Varaiya, P. 2013.
\newblock \emph{The Max-Pressure Controller for Arbitrary Networks of
  Signalized Intersections}, 27--66.
\newblock New York, NY: Springer New York.
\newblock ISBN 978-1-4614-6243-9.
\newblock \doi{10.1007/978-1-4614-6243-9_2}.
\newblock \urlprefix\url{https://doi.org/10.1007/978-1-4614-6243-9_2}.

\bibitem[{Wang et~al.(2018)Wang, Chen, Min, He, Yang, and
  Zhang}]{16wang2018efficient}
Wang, X.; Chen, C.; Min, Y.; He, J.; Yang, B.; and Zhang, Y. 2018.
\newblock Efficient Metropolitan Traffic Prediction Based on Graph Recurrent
  Neural Network.
\newblock ArXiv preprint arXiv:1811.00740.

\bibitem[{Wei et~al.(2019{\natexlab{a}})Wei, Chen, Zheng, Wu, Gayah, Xu, and
  Li}]{4wei2019presslight}
Wei, H.; Chen, C.; Zheng, G.; Wu, K.; Gayah, V.; Xu, K.; and Li, Z.
  2019{\natexlab{a}}.
\newblock PressLight: Learning Max Pressure Control to Coordinate Traffic
  Signals in Arterial Network.
\newblock In \emph{Proceedings of the 25th ACM SIGKDD International Conference
  on Knowledge Discovery and Data Mining (KDD'19)}. Anchorage, AK, USA.

\bibitem[{Wei et~al.(2019{\natexlab{b}})Wei, Xu, Zhang, Zheng, Zang, Chen,
  Zhang, Zhu, Xu, and Li}]{3wei2019colight}
Wei, H.; Xu, N.; Zhang, H.; Zheng, G.; Zang, X.; Chen, C.; Zhang, W.; Zhu, Y.;
  Xu, K.; and Li, Z. 2019{\natexlab{b}}.
\newblock Colight: Learning network-level cooperation for traffic signal
  control.
\newblock In \emph{Proceedings of the 2019 ACM on Conference on Information and
  Knowledge Management (CIKM'19)}. Beijing, China: ACM.

\bibitem[{Wei et~al.(2019{\natexlab{c}})Wei, Zheng, Gayah, and
  Li}]{1wei2019survey}
Wei, H.; Zheng, G.; Gayah, V.; and Li, Z. 2019{\natexlab{c}}.
\newblock A Survey on Traffic Signal Control Methods.
\newblock ArXiv preprint arXiv:1904.08117.

\bibitem[{Xu et~al.(2019)Xu, Dai, Wang, Peng, Xuan, and Guo}]{14xu2019road}
Xu, D.; Dai, H.; Wang, Y.; Peng, P.; Xuan, Q.; and Guo, H. 2019.
\newblock Road traffic state prediction based on a graph embedding recurrent
  neural network under the SCATS.
\newblock \emph{Chaos: An Interdisciplinary Journal of Nonlinear Science}
  29(10): 103125.

\bibitem[{Yu, Yin, and Zhu(2018)}]{19bing2018spatio}
Yu, B.; Yin, H.; and Zhu, Z. 2018.
\newblock Spatio-Temporal Graph Convolutional Networks: A Deep Learning
  Framework for Traffic Forecasting.
\newblock In \emph{Proceedings of the Twenty-Seventh International Joint
  Conference on Artificial Intelligence (IJCAI-18)}. Stockholm, Sweden.

\bibitem[{Zhang et~al.(2019{\natexlab{a}})Zhang, Feng, Liu, Ding, Zhu, Zhou,
  Zhang, Yu, Jin, and Li}]{20ZhangCityFlow}
Zhang, H.; Feng, S.; Liu, C.; Ding, Y.; Zhu, Y.; Zhou, Z.; Zhang, W.; Yu, Y.;
  Jin, H.; and Li, Z. 2019{\natexlab{a}}.
\newblock CityFlow: A Multi-Agent Reinforcement Learning Environment for Large
  Scale City Traffic Scenario.
\newblock ArXiv preprint arXiv:1905.05217.

\bibitem[{Zhang et~al.(2019{\natexlab{b}})Zhang, Wang, Chen, and
  Cao}]{10yuxuan2019gcgan}
Zhang, Y.; Wang, S.; Chen, B.; and Cao, J. 2019{\natexlab{b}}.
\newblock GCGAN: Generative Adversarial Nets with Graph CNN for Network-Scale
  Traffic Prediction.
\newblock In \emph{Proceedings of the International Joint Conference on Neural
  Networks (IJCNN)}. Budapest, Hungary.

\bibitem[{Zhao et~al.(2019)Zhao, Song, Zhang, Liu, Wang, Lin, Deng, and
  Li}]{15ling2019tgcn}
Zhao, L.; Song, Y.; Zhang, C.; Liu, Y.; Wang, P.; Lin, T.; Deng, M.; and Li, H.
  2019.
\newblock T-GCN: A Temporal Graph Convolutional Network for Traffic Prediction.
\newblock \emph{IEEE Transactions on Intelligent Transportation Systems} 1--11.
\newblock Early Access Article.

\bibitem[{Zheng et~al.(2019)Zheng, Xiong, Zang, Feng, Wei, Zhang, Li, Xu, and
  Li}]{2zheng2019learning}
Zheng, G.; Xiong, Y.; Zang, X.; Feng, J.; Wei, H.; Zhang, H.; Li, Y.; Xu, K.;
  and Li, Z. 2019.
\newblock Learning Phase Competition for Traffic Signal Control.
\newblock In \emph{Proceedings of the 2019 ACM on Conference on Information and
  Knowledge Management (CIKM'19)}. Beijing, China: ACM.

\bibitem[{Zhiyong et~al.(2019)Zhiyong, Kristian, Ruimin, and
  Yinhai}]{7zhiyong2019traffic}
Zhiyong, C.; Kristian, H.; Ruimin, K.; and Yinhai, W. 2019.
\newblock Traffic Graph Convolutional Recurrent Neural Network: A Deep Learning
  Framework for Network-Scale Traffic Learning and Forecasting.
\newblock \emph{IEEE Transactions on Intelligent Transportation Systems} 1--12.
\newblock Early Access Article.

\bibitem[{Zhou et~al.(2020)Zhou, Yang, Zhang, Trajcevski, Zhong, and
  Khokhar}]{18zhou2020reinforced}
Zhou, F.; Yang, Q.; Zhang, K.; Trajcevski, G.; Zhong, T.; and Khokhar, A. 2020.
\newblock Reinforced Spatio-Temporal Attentive Graph Neural Networks for
  Traffic Forecasting.
\newblock \emph{IEEE Internet of Things Journal} 1--16.
\newblock Early Access Article.

\end{thebibliography}

\end{document}